\documentclass[sigconf]{acmart}

\AtBeginDocument{%
  \providecommand\BibTeX{{%
    \normalfont B\kern-0.5em{\scshape i\kern-0.25em b}\kern-0.8em\TeX}}}

\setcopyright{acmcopyright}
\copyrightyear{2019}
\acmYear{2019}
\begin{document}

\title{WiredSwarm: High Resolution Haptic Feedback Provided by a Swarm of Drones to the User's Fingers for VR interaction}

\author{Evgeny Tsykunov}
\email{Evgeny.Tsykunov@skoltech.ru}
\affiliation{%
  \institution{Skolkovo Institute of Science and Technology}
  \city{Moscow}
  \country{Russia}
}

\author{Dzmitry Tsetserukou}
\email{D.Tsetserukou@skoltech.ru}
\affiliation{%
  \institution{Skolkovo Institute of Science and Technology}
  \city{Moscow}
  \country{Russia}
}

\renewcommand{\shortauthors}{Evgeny Tsykunov, et al.}

\begin{abstract}

We propose a concept of a novel interaction strategy for providing rich haptic feedback in Virtual Reality (VR), when each user's finger is connected to micro-quadrotor with a wire. Described technology represents the first flying wearable haptic interface. The solution potentially is able to deliver high resolution force feedback to each finger during fine motor interaction in VR. The tips of tethers are connected to the centers of quadcopters under their bottom. Therefore, flight stability is increasing and the interaction forces are becoming stronger which allows to use smaller drones.

\end{abstract}





\maketitle

\section{Introduction}
A noticeable part of human-robot interaction involves the usage of haptic and tactile feedback \cite{Akahane_2013, Tsykunov_2019}.
Along with that, there are several research projects related to  providing encountered-type kinesthetic and tactile feedback in VR via drones.

In \cite{vrhapticdrones}, an object or surface is connected to the drone which is supposed to be touched by a human to deliver passive or active tactile feedback. Authors in \cite{tactile_drones} also propose to hit the user with some object connected to a small drone to provide a haptic sensation. Abdullah et al. in \cite{hapticdrone} also proposed to push or pull the drone in $Z-axis$ direction to simulate force feedback for direct interaction. 
Abtahi et al. in their paper \cite{beyond_the_force} developed more complicated scenario which incorporates rich interactions including passive force feedback and texture mapping. 

The main limitations of the proposed solutions include low sensation resolution, instability during interaction, low impact force, and big size of a drone. In particular, authors in \cite{hapticdrone, beyond_the_force} selected bigger size drones. Usually spacial motion of human hands is fast. Although more powerful quadrotors could provide more noticeable force feedback, they could be slow for certain application. It is also hard to combine different types of feedback at the same time because there are not enough space for drones near the fingers. 

\begin{figure}[t]
\centering
\includegraphics[width=0.49\textwidth]{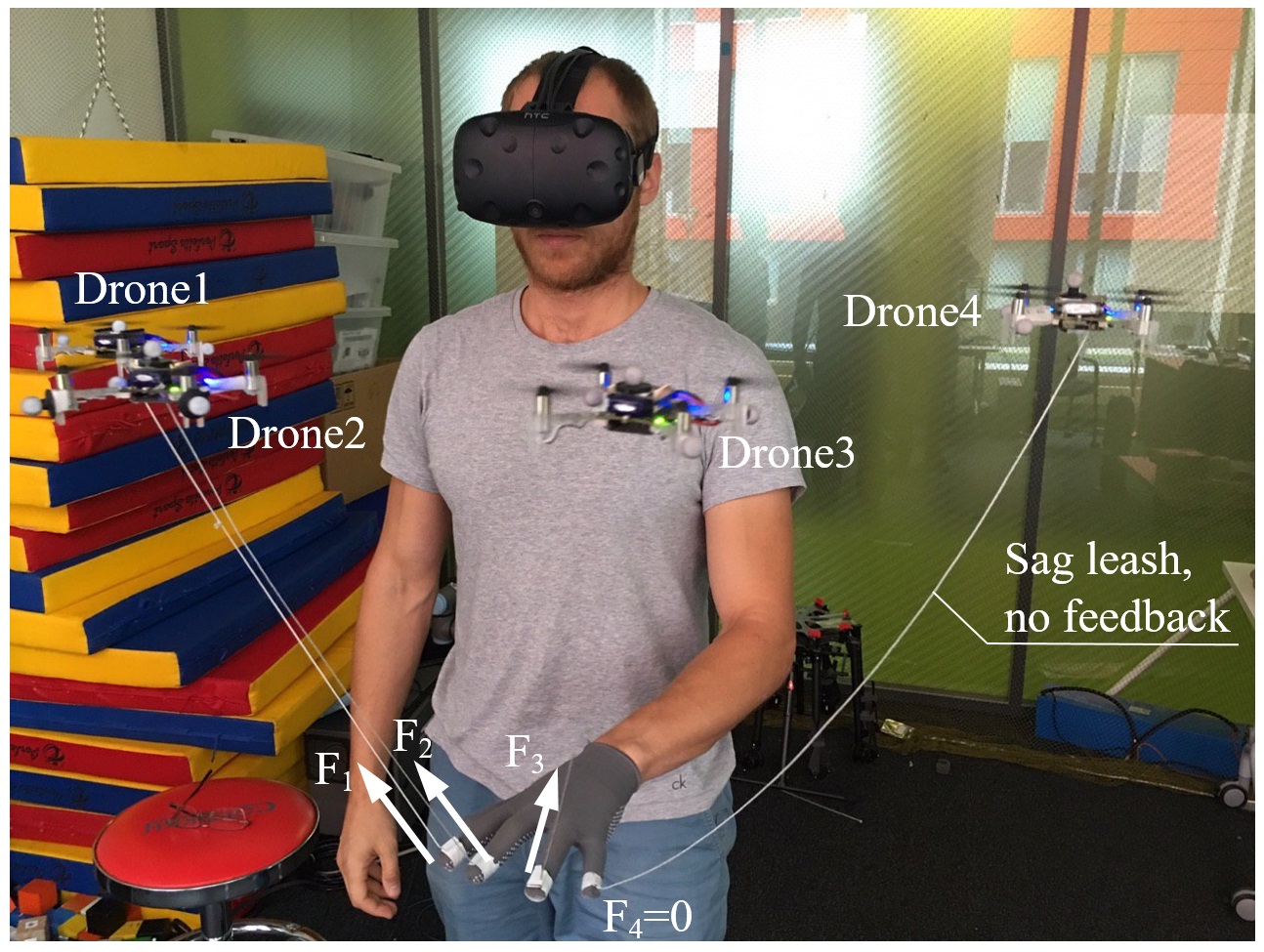}\label{true}
\caption{Interaction of a user with a virtual scene using flying wearable haptic interface WiredSwarm.}
\label{first}
\end{figure}

%   Related Work

%     I would recommend the authors to consider two VR papers related to haptics that
%     were published recently:
%     --Evaluation of presence in virtual environments: haptic vest and user's haptic
%     skills
%     --Human–virtual character interaction: Toward understanding the influence of
%     haptic feedback

\section{Main contributions}
In contrast with the discussed works, we propose to connect multiple micro-quadrotors and each finger with leashes. String is able to deliver a force vector (direction and magnitude) directly from the drone to the finger.

\subsection{High sensation resolution}
Due to the leash length, the drones are not in the close proximity near the fingers, therefore we can use multiple of them to deliver haptic feedback to the close regions - neighbour fingers (shown in Fig. \ref{first}). One option is to use one drone to deliver force to each finger. Another way is to combine five fingers into three groups and connect one drone to each group.

Two drones with different tether properties can be connected to the same finger.
Combination of different types of leash (e.g., elastic or non-stretchable) could deliver a rich kinaesthetic  feedback.

Using more that one drone allows to create high resolution encountered-type haptic feedback. In this case each finger has to be tracked by a localization system, such as Leap Motion.

\subsection{Flight stability}
During the interaction session the drones are not visible in VR, which rises additional safety challenges. 
Usual physical interaction strategy involves direct contact between the fingers and some part of the drone. The point of contact is far from the center of the drone \cite{vrhapticdrones, beyond_the_force}, which generates high and sudden external moment. Such a high moment leads to the instabilities which are hard to overcome by a flight controller.

In the proposed solution tether is connected to the center of the drone under the bottom providing higher flight stability. 
The suggested design implies the selection of leash properties. One option is to make an elastic leash, which ensures smooth interaction and more safe flight.

\subsection{Higher impact force}
Each rotor, rotating at speed $\omega_i$ produces a force  and a momentum which are defined as:
\begin{equation}
    F_i=k_f \omega_i^2, M_i=k_m\omega_i^2
\end{equation}
where $k_f$ is the force coefficient, $k_m$ is the momentum coefficient, $i=1,2,3,4$ for the quadrotor \cite{Corke}.
The resulting control inputs to the drone frame are expressed as:

\begin{equation} \label{eq:control_input}
    \textbf{u} 
    = 
    \begin{bmatrix} u_1 \\ u_2 \\ u_3 \\ u_4 \end{bmatrix}
    =
    \begin{bmatrix}
    k_f & k_f & k_f & k_f \\
    0 & k_f L & 0 & -k_f L \\
    -k_f L & 0 & k_f L & 0 \\
    k_m & k_m & k_m & k_m    
    \end{bmatrix}
    \begin{bmatrix} \omega_1^2 \\ \omega_2^2 \\ \omega_3^2 \\ \omega_4^2 \end{bmatrix}
\end{equation}

The coordinate systems of the drone and world frame are defined as $D$ and $W$, respectively.
The majority of conducted research \cite{vrhapticdrones,beyond_the_force,tactile_drones} utilize force $u_1$ that is produced by the rotors in $x_W$ or $y_W$ direction. However, $u_1$ acts along $z_D$ axis.
During the interaction with human, drones usually avoid highly aggressive maneuvers with high acceleration, pitch, and roll angles, which produces small $z_D$ inclination. Force calculated as $u_1 \textbf{z}_D$ has a small horizontal term (along $x_W$ and $y_W$ axis), and, therefore, the impact with human is negligible.

Authors in \cite{beyond_the_force} also partially employ the moments $u_2$ and $u_3$ for the surface stiffness simulation. $u_2$ defined as:
\begin{equation} \label{eq:moment}
    u_2 = k_f \omega_2^2 L - k_f \omega_4^2 L
\end{equation}
It is possible to see from \eqref{eq:moment} that the potential impact from moment (combination of $u_2$ and $u_3$) is limited, comparing to $u_1$. Only up to two rotors actually contribute to the force feedback. In addition, $u_1$ which requires a lot of power, suffers from highlighting $u_2$ and $u_3$.

We propose the concept in which the tether is connected to the bottom of the quadrotor and the main force direction is along $z_w$ axis. Therefore, the most powerful control input $u_1$ is used to deliver the force feedback while the drone is in well-stabilized position.

\section{System implementation}
We used Crazyflie 2.0 quadrotors and Robot Operating System (ROS) environment for the drone control.
The state of the human hand and drones (position and orientation) are estimated with a Vicon motion capture system (12 Vantage V5 cameras are covering $5m \times  5 m \times  5 m$ space). The position and attitude update rate was 100 Hz for all drones.
For easy setup, we propose to connect the tethers to the fingers with magnets.

Drones follow the human hand while the user is moving in the virtual scene.
We propose to predict the intersection between the human hand and the virtual object and the corresponding drones become activated and provide fine force feedback.

\section{Conclusion and Future Work}
The novel type of device, flying wearable glove, was developed by us. This type of the haptic interface considerably increases the working area comparing with the desktop type devices. Additionally, it does not produce fatigue to the user, as it is often the case with wearable devices, because of zero weight.

Specifically, the proposed technology provides strong advantages for rendering the interaction with  horizontal virtual surfaces. For example, playing virtual musical instruments, e.g., virtual piano can be a scenario for its application. Moreover, application of WiredSwarm in VR surgical simulator will allow the user to feel the interaction with tissue, muscles, and bones.  

WiredSwarm technology can be extended to deliver haptic sensation not only to the fingers but also to forearms and upper arms. To achieve rendering of textures of virtual objects it is reasonable to integrate vibromotors at the lower tip of the strings. In the future work we will add experimental data and theoretical analysis on the stability, resolution of force feedback, validation in a VR scenario.

% \begin{acks}
% \end{acks}

\bibliographystyle{ACM-Reference-Format}
\bibliography{sample-base}

% \appendix









\end{document}